\pdfoutput=1

\documentclass[11pt]{article}

\usepackage[final]{acl}

\usepackage{amsmath,amsfonts,amssymb}
\usepackage{bm}
\usepackage{mathtools}
\usepackage{physics}
\usepackage{ascmac}
\usepackage{amsthm}
\usepackage{float}
\usepackage{listings}
\restylefloat{table}
\usepackage{xurl}
\usepackage{times}
\usepackage{latexsym}

\usepackage[T1]{fontenc}

\usepackage[utf8]{inputenc}

\usepackage{microtype}

\usepackage{inconsolata}
\usepackage{booktabs}
\usepackage{multirow}

\title{NeuronMoE: Neuron-Guided Mixture-of-Experts for Efficient \\Multilingual LLM Extension}

\author{  Rongzhi Li\textsuperscript{1,2} \quad Hitomi Yanaka\textsuperscript{1,2,3}\\
  \textsuperscript{1}The University of Tokyo \quad \textsuperscript{2}Riken \quad \textsuperscript{3}Tohoku University\\
  \texttt{\{iimori-eiji, hyanaka\}@is.s.u-tokyo.ac.jp} \\}

\begin{document}
\maketitle
\begin{abstract}
Extending large language models to low-resource languages is essential for global accessibility, but training separate models per language is prohibitively expensive.
Mixture-of-Experts (MoE) architectures address this by adding sparse language-specific parameters, but determining how many experts each layer needs remains an open question.
Current approaches allocate experts based on layer-level similarity, yet language processing exhibits fine-grained specialization at individual neurons.
We propose \textbf{NeuronMoE}, a method that analyzes language-specific neurons across all transformer components to guide expert allocation per layer based on empirically measured cross-lingual neuron diversity.
Applied to Llama-3.2-3B for low-resource languages (Greek, Turkish, and Hungarian), this approach achieves approximately 40\% average parameter reduction  while matching the performance of the LayerMoE baseline. 
We find that low-resource language experts independently develop neuron specialization patterns mirroring the high-resource language, which are concentrated in early and late layers. This reveals potential universal architectural principles in how multilingual models organize linguistic knowledge.
Our approach generalizes across architectures, as validated on Qwen, and shows that allocation strategy matters more than total expert count.\footnote{Our code is available at \url{https://github.com/ynklab/NeuronMoE}.} 
\end{abstract}

\section{Introduction}

Extending Large Language Models (LLMs) to support multiple languages is essential for global information access. However, building high-performance models for low-resource languages remains challenging due to data scarcity and computational costs \citep{joshi2020state}.
Mixture-of-Experts (MoE) architectures offer a promising solution through sparse expert activation, enabling efficient model scaling.
Recent work has applied MoE to multilingual and low-resource language extension \citep{li-etal-2023-mmnmt,zhu-etal-2024-llama,zhou2025moe,li-etal-2025-group}, with layer-wise allocation strategies \citep{zhang-etal-2025-less} achieving parameter reduction by allocating experts based on cross-lingual similarity.
However, similarity-based allocation provides only an indirect signal of language-specific processing requirements, as high similarity does not necessarily mean low capacity needs.
They also rely on attention-layer similarity while ignoring MLP layers that comprise two-thirds of model parameters.

\begin{figure}[t]
\centering
\includegraphics[width=0.48\textwidth]{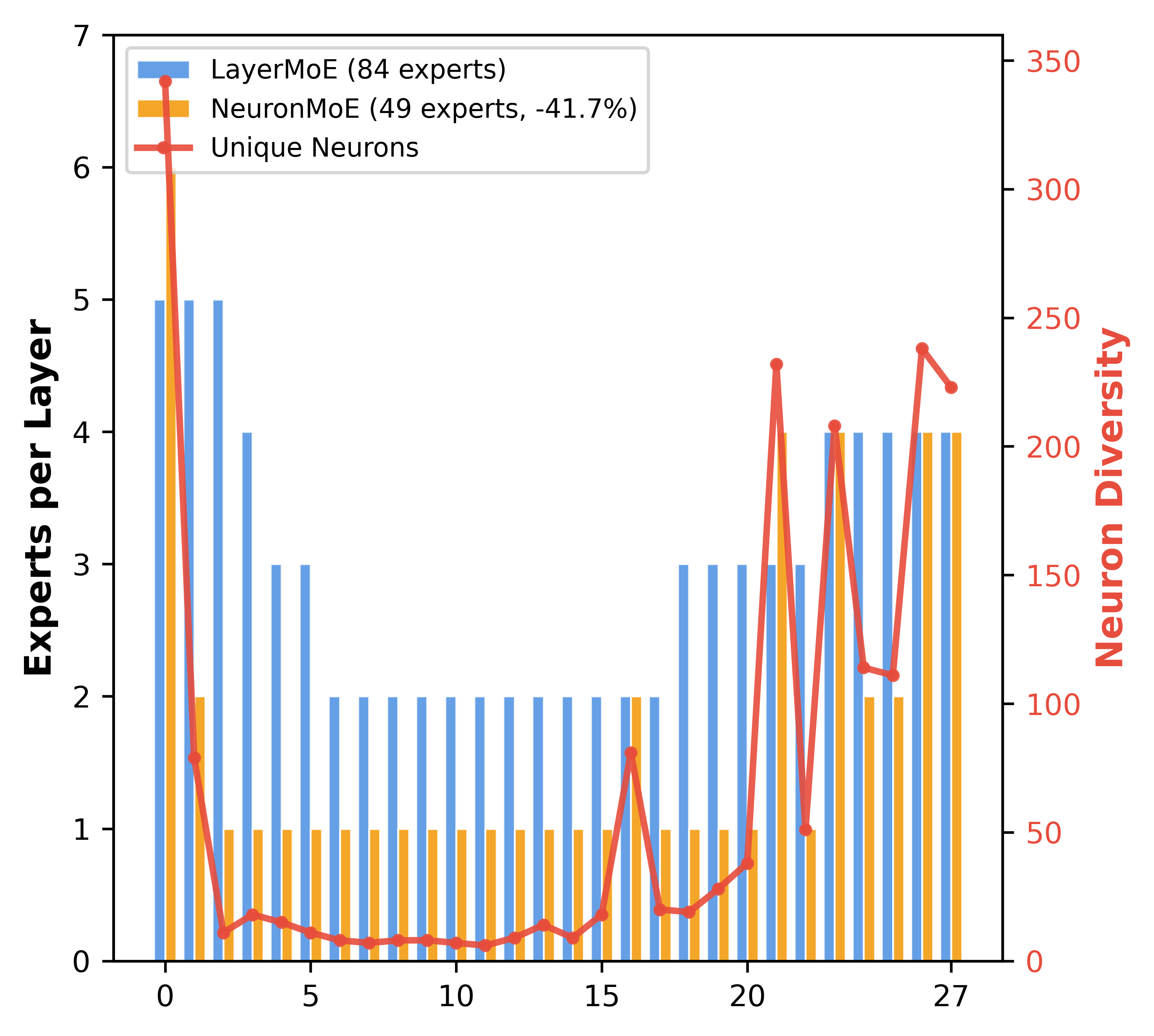}
\caption{Expert allocation comparison on Llama-3.2-3B for Greek extension. Cross-lingual neuron diversity (red line) exhibits heterogeneous distribution across layers. NeuronMoE allocates experts following this distribution, achieving 41.7\% parameter reduction (49 vs 84 experts) compared to LayerMoE's uniform allocation strategy.}
\end{figure}

On the other hand, studies on language-specific neurons \citep{durrani-etal-2020-analyzing,tang-etal-2024-language,kojima-etal-2024-multilingual} reveal that individual neurons encode language knowledge, with heterogeneous distribution across layers, which is concentrated in early and late layers and sparse in middle layers. 
However, their practical application mostly remains limited to observational studies. 

\begin{figure*}[t]
\centering
\includegraphics[width=0.9\textwidth]{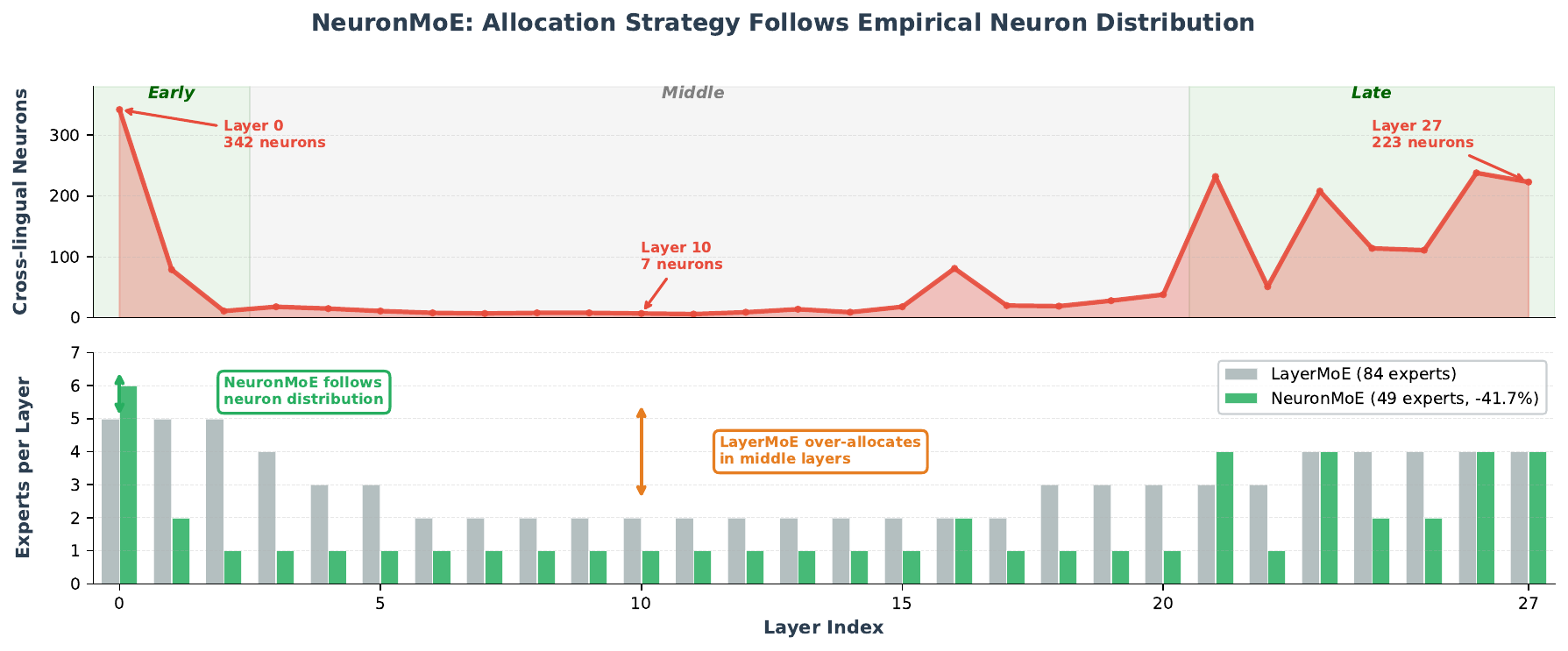}
\caption{Overview of NeuronMoE's core insight and allocation strategy. \textbf{Top:} Cross-lingual neuron diversity exhibits dramatic heterogeneity across layers. For example, Layer 0 contains 342 unique language-specific neurons while Layer 10 contains only 7, demonstrating that different layers have vastly different language-specific processing requirements. \textbf{Bottom:} NeuronMoE allocates experts following this empirical distribution (49 experts total), concentrating capacity where specialization occurs. In contrast, LayerMoE's similarity-based approach over-allocates in middle layers (84 experts total). This neuron-guided strategy achieves 41.7\% parameter reduction while maintaining comparable performance.}
\label{fig:overview}
\end{figure*}

This work bridges these two lines of research by using neuron-level language specialization analysis to guide MoE expert allocation.
Figure~\ref{fig:overview} illustrates our approach.
We analyze language-specific neurons across all transformer components (both attention and MLP layers) and then allocate experts based on empirically measured specialization needs.

We investigate two questions: (1) \textbf{Can neuron-level analysis enable more efficient expert allocation than layer-level similarity?}(2)\textbf{ How do language-specific neurons develop within MoE experts during multilingual extension?}

To address (1), our experiments demonstrate that neuron-guided allocation on Llama-3.2-3B \citep{grattafiori2024llama3herdmodels} achieves 42\% average parameter reduction (47-50 vs. 84 experts) while maintaining comparable performance across different low-resource languages (Greek, Turkish, and Hungarian).
By directly measuring language-specific capacity requirements rather than relying on similarity-based allocation, our method concentrates experts in layers with empirically identified high specialization needs.
This efficiency generalizes across typologically diverse language families including Indo-European, Turkic, and Uralic. Validation on Qwen-1.5-1.8B \cite{qwen1.5} achieves 50\% reduction, confirming that allocation strategy determines parameter efficiency.

Regarding (2), our analysis reveals that target language experts independently develop neuron specialization patterns mirroring those of the source language, whose neurons concentrate in early and late layers despite typological differences, indicating universal architectural principles governing how multilingual models organize linguistic knowledge.\footnote{Code is available at \url{https://github.com/ynklab/NeuronMoE}}

\section{Related Work}

\subsection{Multilingual Extension via Mixture-of-Experts}

Extending pre-trained LLMs to support additional languages presents a fundamental challenge: catastrophic forgetting of original language capabilities during acquisition of new language knowledge.
Mixture-of-Experts (MoE) architectures address this through sparse expert activation, enabling language-specific adaptation while preserving original capabilities.

Early MoE applications to multilingual scenarios employed uniform expert allocation, assigning identical expert counts across all layers \citep{lepikhin2020gshard,fedus2021switch,du2022glam}.
\citet{komatsuzaki2022sparse} propose sparse upcycling, converting pre-trained dense models into MoE architectures by initializing experts from dense checkpoints.
While these uniform allocation strategies demonstrate strong scaling properties, they do not exploit layer-specific optimization opportunities based on varying linguistic processing requirements across layers.

\citet{zhou2025moe} introduce MoE-LPR (Language Priors Routing), a two-stage training framework for multilingual extension.
Stage 1 freezes original model parameters while training newly added experts on target language data, preserving base model knowledge.
Stage 2 trains routing mechanisms with minimal replay data (less than 1\% of Stage 1 data), incorporating language priors to recover original capabilities.
While this framework effectively mitigates catastrophic forgetting, it maintains uniform expert allocation across all layers.

Building on MoE-LPR, \citet{zhang-etal-2025-less} introduce layer-wise allocation based on cross-lingual similarity (hereafter, we call this framework LayerMoE).
They allocate experts inversely proportional to attention-layer similarity: layers with high cross-lingual similarity receive fewer experts, while layers with low similarity receive more experts.
This strategy achieves 60\% parameter reduction over uniform allocation in single-language extension and 33.3\% in continual multilingual scenarios.

However, their approach faces limitations that motivate our work.
Their similarity computation considers only attention layers, omitting MLP layers that comprise two-thirds of model parameters and encode substantial linguistic knowledge \citep{durrani-etal-2020-analyzing}.
Layer-level similarity provides coarse allocation signals that may not capture fine-grained language-specific processing requirements within layers.
Moreover, their similarity-based allocation strategy lacks grounding in empirically measured language specialization patterns identified based on neuron-level analysis.

We address these limitations through neuron-level language specialization analysis across all transformer components, providing more direct and comprehensive signals for expert allocation decisions.

\subsection{Language-Specific Neurons in Multilingual Models}

A parallel line of research investigates how multilingual models encode language-specific knowledge at the neuron level, revealing fine-grained specialization patterns within model architectures.

\citet{durrani-etal-2020-analyzing} pioneer neuron-level analysis of pre-trained models, demonstrating that core linguistic properties can be predicted from small neuron subsets, suggesting concentrated language encoding.
Recent work extends this to decoder-only LLMs with varying methodological approaches for identifying language-specific neurons.

\paragraph{Identification Methods:}
\citet{tang-etal-2024-language} propose Language Activation Probability Entropy (LAPE) for MLP layers, measuring activation probability entropy across language-specific inputs. Neurons exhibiting low entropy for specific languages are identified as language-specific.
\citet{kojima-etal-2024-multilingual} define language-specific neurons as those exhibiting statistically significant activation patterns for individual languages across both attention and MLP layers, measuring significance through Average Precision scores.
Their analysis reveals minimal overlap between languages, with each language developing distinct specialized neuron populations.

\paragraph{Layer-wise Heterogeneity:}
Despite methodological differences, these works converge on a critical observation: language specialization exhibits heterogeneous distribution across layers.
\citet{tang-etal-2024-language,kojima-etal-2024-multilingual} demonstrate that language-specific neurons concentrate in early layers and late layers, while middle layers contain minimal language-specific neurons, suggesting abstract, language-agnostic processing.
This heterogeneous pattern has implications for architectural design: layers with high language specialization may benefit from multiple experts to accommodate diverse linguistic patterns, while language-agnostic layers may require fewer parameters.

\paragraph{Current Applications and Limitations:}
Recent work explores multilingual activation patterns \citep{liu-etal-2024-unraveling} and linguistic neuron overlap for cross-lingual transfer \citep{chen2025linguistic,xu2025linguisticneuronoverlappatterns}, though \citet{mondal-etal-2025-language} question their utility for transfer.
Practical applications have been limited to observational studies, model interpretation, and output manipulation.
Leveraging neuron-level signals to guide architectural design decisions, such as expert allocation in MoE, remains unexplored.

\paragraph{Bridging Neuron Analysis and MoE Allocation:}
The observed heterogeneous distribution of language-specific neurons provides a direct signal for layer-specific expert allocation.
Unlike layer-wise similarity, which indirectly infers capacity needs by measuring representation alignment, neuron counting directly quantifies how many specialized units each layer requires for multilingual processing.
This key insight motivates our approach: if a layer contains 342 unique language-specific neurons across two languages while another contains only 7, the former empirically requires greater capacity for language-specific computation.
By measuring this distribution before MoE training, we can allocate experts proportionally to empirically observed specialization rather than relying on similarity-based allocation.

\section{Method}
We bridge neuron-level analysis and MoE architecture design by using empirically measured language specialization patterns to guide expert allocation.
Our framework is compatible with any neuron specificity metric that quantifies language-specific activation patterns.
In this work, we adopt \citet{kojima-etal-2024-multilingual}'s Average Precision method for its comprehensive coverage of both attention and MLP layers; alternative metrics such as LAPE \citep{tang-etal-2024-language} could also be applied.
Rather than relying on indirect measures such as attention-layer similarity, we directly measure cross-lingual neuron diversity to determine layer-specific expert requirements.

\subsection{Neuron Specialization Measurement}

\paragraph{Language-Specific Neuron Definition:}
Following \citet{kojima-etal-2024-multilingual}, we define language-specific neurons as neurons that exhibit statistically significant activation patterns for individual languages.
For a given neuron $n$ in layer $l$ and language $\ell$, we compute activation values $\{a_n^{(i)}\}$ across a corpus of samples $\{x^{(i)}\}$, where $x^{(i)}$ is labeled with language $y^{(i)} \in \{\ell_1, \ldots, \ell_K\}$.

The language specificity of a neuron $n$ for language $\ell$ is measured using Average Precision (AP). AP represents the degree to which samples of language $\ell$ are concentrated at the top when sorted by activation values in descending order:
\begin{multline}
\text{AP}(n, \ell) = \frac{1}{|\{i : y^{(i)} = \ell\}|} \\
\sum_{k=1}^{N} \text{P}(k) \cdot \mathbf{1}[y^{(\pi(k))} = \ell]
\end{multline}
where $\pi$ is the descending sort of activation values, and $\text{P}(k)$ is the proportion of language $\ell$ in the top $k$ items. A neuron is considered more specific to language $\ell$ as AP approaches 1.0.

\paragraph{Cross-Lingual Neuron Diversity for Expert Allocation:}
Following \citet{kojima-etal-2024-multilingual}, we identify the top 1000 language-specific neurons per language across all layers based on their AP scores, using AP $> 0.5$ as the threshold.
We then determine expert allocation based on the diversity of language-specific neurons within each layer across the source language and a single target language.
The layer specialization score $\mathcal{S}_l$ is defined as the total number of unique language-specific neurons in layer $l$, aggregated across the language pair.
For Greek extension, this is formulated as:

\begin{equation}
\mathcal{S}_l = \left| \bigcup_{\text{lang} \in \{\text{en, el}\}} N_{l, \text{lang}} \right|
\end{equation}

where $N_{l, \text{lang}}$ is the set of language-specific neurons for layer $l$ and language lang.
This score reflects the total pool of specialized neurons a layer needs to handle both languages.
The same formulation applies to Turkish (en, tr) and Hungarian (en, hu) extensions.

\subsection{Expert Allocation Strategy}
The number of experts for each layer, $E_l$, is determined by linearly scaling the unique neuron count $\mathcal{S}_l$ between a predefined minimum and maximum number of experts.

First, the unique neuron counts are normalized to a [0, 1] range:
\begin{equation}
\text{norm}(\mathcal{S}_l) = \frac{\mathcal{S}_l - \min(\mathcal{S})}{\max(\mathcal{S}) - \min(\mathcal{S})}
\end{equation}

Then, the number of experts is calculated as:
\begin{equation}
E_l = E_{\min} + \text{round}(\text{norm}(\mathcal{S}_l) \times (E_{\max} - E_{\min}))
\end{equation}

where $E_{\min}$ and $E_{\max}$ are the minimum and maximum allowed experts per layer.
This data-driven approach allocates more experts to layers with a greater diversity of language-specific neurons.

\subsection{Two-Stage Training Process}
Following the MoE-LPR framework \citep{zhou2025moe}, we adopt a two-stage training approach where our neuron-guided allocation replaces the similarity-based allocation for determining expert counts per layer.

\paragraph{Stage 1 (Expert Initialization):} We freeze the original pre-trained model parameters and add new MoE experts to each layer according to our neuron-guided allocation strategy.
The model is then trained on target language data (Greek, Turkish, or Hungarian) to initialize these experts.
At this stage, expert counts are determined by the layer specialization scores $\mathcal{S}_l$ computed from cross-lingual neuron diversity while LayerMoE \citep{zhang-etal-2025-less} uses attention-layer similarities for the expert allocation.
This stage focuses on adapting experts to the new language while preserving original parameters.

\paragraph{Stage 2 (Router Training):} Following LayerMoE \citep{zhang-etal-2025-less}, we train the routing mechanism using a small amount of replay data from the source languages mixed with target language data, incorporating language priors to recover original language capabilities and refine expert selection.
The key difference between our method and LayerMoE lies solely in Stage 1's allocation strategy; Stage 2 routing training is identical for both methods.

\section{Experimental Setup}

To enable direct comparison with LayerMoE \citep{zhang-etal-2025-less}, we adopt their experimental setup: Llama-3.2-3B (28 layers) \citep{grattafiori2024llama3herdmodels} as our primary model and Qwen-1.5-1.8B (24 layers) \citep{qwen1.5} for cross-architecture validation.
Following their language selection, we extend these models to Greek (EL), Turkish (TR), and Hungarian (HU) using 2B tokens per language from CulturaX. These languages represent diverse families: Indo-European, Turkic, and Uralic.
We evaluate using the lm-evaluation-harness framework \citep{eval-harness}, which provides multilingual versions of ARC Challenge, MMLU, and HellaSwag translated by Okapi \citep{lai-etal-2023-okapi}. We also use Belebele \citep{bandarkar-etal-2024-belebele} for reading comprehension. Evaluation settings are 5-shot for MMLU, 25-shot for ARC Challenge, and 0-shot for HellaSwag and Belebele.
All results are reported as single-run evaluations.

We conduct experiments across three target languages and two model architectures:

\textbf{Greek extension:} Full evaluation on Llama-3.2-3B covering both training stages, plus cross-architecture validation on Qwen-1.5-1.8B (Stage 1 only). Baselines include Dense (no MoE), LayerMoE-S1/LayerMoE (84 experts for Llama, 72 for Qwen, using attention-layer similarity allocation \citep{zhang-etal-2025-less}), and NeuronMoE variants (49 experts for Llama, 36 for Qwen). We include an ablation NeuronMoE-EN (37 experts) that allocates based solely on English neuron distributions, testing whether cross-lingual neuron analysis is essential.

\textbf{Turkish and Hungarian extensions:} Validation experiments on Llama-3.2-3B at Stage 1 only, comparing Dense, LayerMoE (84 experts), and NeuronMoE (50 and 47 experts respectively) to assess cross-lingual generalization across typologically diverse language families (Indo-European, Turkic, Uralic).

All models are trained using the AdamW optimizer with learning rate 1e-4, cosine scheduler, batch size 32, and 15K training steps.
Full hyperparameters are detailed in Appendix \ref{sec:training}.
The neuron analysis preprocessing requires approximately 6 minutes per language on a single NVIDIA GH200 GPU.

\section{Results}

\subsection{Greek Extension on Llama-3.2-3B}

\begin{table*}[t]
\centering
\small
\begin{tabular}{lrrrrrrrrr}
\toprule
& & \multicolumn{2}{c}{\textbf{ARC}} & \multicolumn{2}{c}{\textbf{Belebele}} & \multicolumn{2}{c}{\textbf{HellaSwag}} & \multicolumn{2}{c}{\textbf{MMLU}} \\
\cmidrule(lr){3-4} \cmidrule(lr){5-6} \cmidrule(lr){7-8} \cmidrule(lr){9-10}
\textbf{Model} & \textbf{\#Exp} & \textbf{EN} & \textbf{EL} & \textbf{EN} & \textbf{EL} & \textbf{EN} & \textbf{EL} & \textbf{EN} & \textbf{EL} \\
\midrule
\multicolumn{10}{l}{\textit{Llama-3.2-3B (28 layers):}} \\
Dense & - & 51.11 & 31.93 & 74.11 & 65.22 & 76.33 & 43.38 & 56.45 & 41.17 \\
\midrule
\multicolumn{10}{l}{\quad \textit{Stage 1 (Expert Initialization):}} \\
\quad LayerMoE-S1 & 84 & 47.27 & 35.19 & 73.89 & 66.33 & 75.84 & 52.58 & 55.40 & 43.41 \\
\quad NeuronMoE-EN & 37 (-56\%) & 49.15 & 33.13 & 73.56 & 63.67 & 76.45 & 49.03 & 56.89 & 43.26 \\
\quad NeuronMoE-S1 & 49 (-42\%) & 49.15 & 34.16 & 75.33 & 63.56 & 76.34 & 50.69 & 56.61 & 43.95 \\
\midrule
\multicolumn{10}{l}{\quad \textit{Stage 2 (Routing Training):}} \\
\quad LayerMoE & 84 & 49.32 & 37.50 & 74.78 & 67.33 & 76.50 & 52.41 & 55.79 & 44.06 \\
\quad NeuronMoE & 49 (-42\%) & 50.17 & 35.02 & \
75.11 & 64.56 & 76.53 & 50.53 & 56.48 & 43.66 \\
\midrule
\multicolumn{10}{l}{\textit{Qwen-1.5-1.8B (24 layers, Stage 1 only):}} \\
\quad LayerMoE-S1 & 72 & 37.29 & 25.26 & 54.89 & 31.44 & 59.88 & 35.21 & 45.46 & 31.26 \\
\quad NeuronMoE-S1 & 36 (-50\%) & 37.88 & 23.80 & 58.67 & 32.00 & 61.20 & 31.26 & 45.46 & 29.23 \\
\bottomrule
\end{tabular}
\caption{Performance on Greek extension across Llama-3.2-3B and Qwen-1.5-1.8B. \#Exp indicates the number of experts. NeuronMoE achieves 41.7-50.0\% parameter reduction with comparable performance, demonstrating cross-architecture generalization.}
\label{tab:main}
\end{table*}

Table~\ref{tab:main} presents results for Greek extension on Llama-3.2-3B across both training stages.
NeuronMoE-S1 achieves 41.7\% parameter reduction with 49 experts compared to LayerMoE-S1's 84, while mitigating catastrophic forgetting in English with consistent +0.5-2\% gains across benchmarks and maintaining comparable Greek performance.
After Stage 2 routing training, NeuronMoE maintains near-Dense English performance while providing substantial Greek capability (+3.09\% over Dense baseline on ARC), demonstrating successful multilingual extension without sacrificing original language proficiency.

NeuronMoE achieves 41.7\% parameter reduction at the cost of 1-2.8\% target language performance on specific benchmarks. The tradeoffs are task-dependent: ARC Challenge shows 2.0-2.5\% degradation, while language understanding tasks such as Belebele, HellaSwag, and MMLU show smaller gaps of 0.4-2.8\%.
This pattern reflects our allocation strategy that concentrates experts in early/late layers handling language-specific processing while minimizing allocation in middle layers performing abstract reasoning.

\subsection{Ablation Study: Importance of Target Language Analysis}

The NeuronMoE-EN ablation allocates 37 experts using only English neuron distribution. Despite achieving strong English performance at 49.15\% on ARC, Greek performance drops to 33.13\% compared to 34.16\% for full NeuronMoE.
This confirms that effective MoE allocation requires analyzing neuron specialization patterns for the target language, not just source language patterns.

\subsection{Cross-Architecture Validation}

Qwen-1.5-1.8B results in Table~\ref{tab:main} demonstrate that neuron-guided allocation generalizes across architectures.
Despite architectural differences between the 24-layer Qwen and 28-layer Llama with different hidden dimensions, NeuronMoE achieves 50.0\% parameter reduction.
At Stage 1, Qwen NeuronMoE-S1 shows strong English improvements over LayerMoE-S1 on Belebele (+3.78\%) and HellaSwag (+1.32\%), paralleling Llama results.
Greek ARC performance decreases 1.46\% compared to LayerMoE-S1, suggesting that while neuron-level specialization patterns reflect universal principles, smaller models may exhibit different sensitivities for reasoning tasks.

\subsection{Cross-Lingual Generalization}
To evaluate cross-lingual generalization, we extend Llama-3.2-3B to Turkish and Hungarian using the same neuron-guided allocation strategy (50 and 47 experts respectively).
Table~\ref{tab:additional_langs} shows that NeuronMoE successfully extends to both languages while maintaining approximately 2\% ARC degradation compared to LayerMoE (84 experts), consistent with the Greek results.
Performance remains substantially above Dense baseline (+0.94\% for Turkish, +2.82\% for Hungarian on ARC).
The consistent performance-efficiency tradeoff pattern across three typologically diverse language families (Indo-European, Turkic, Uralic) validates cross-lingual generalization.

\subsection{Analysis of Allocation Strategy}

Table~\ref{tab:neuron_diversity} in Appendix~\ref{sec:neuron_distribution} shows cross-lingual neuron diversity across layers. Early layers exhibit high neuron counts, with L0 containing 342 unique neurons for EN+EL. Late layers show moderate to high counts ranging from 111 to 238 neurons at L21 and L23-27. Middle layers contain minimal specialization, with L7 and L10 having only 7 neurons each.

\begin{table}[t]
\centering
\small
\begin{tabular}{lcc}
\toprule
\textbf{Layers} & \textbf{LayerMoE} & \textbf{NeuronMoE} \\
\midrule
0-2 (Early) & 5,5,5 & 6,2,1 \\
3-10 (Early-Mid) & 4,3,3,2,2,2,2,2 & 1,1,1,1,1,1,1,1 \\
11-18 (Mid-Late) & 2,2,2,2,2,2,2,3 & 1,1,1,1,1,1,1,2 \\
19-27 (Late) & 3,3,3,3,4,4,4,4,4 & 1,1,4,1,4,2,2,4,4 \\
\midrule
\textbf{Total} & \textbf{84} & \textbf{49} \\
\bottomrule
\end{tabular}
\caption{Layer-wise expert allocation comparison (Llama-3.2-3B, Greek extension).
NeuronMoE achieves 41.7\% parameter reduction by concentrating experts in high-diversity layers while minimizing allocation in middle layers.}
\label{tab:allocation}
\end{table}

\begin{table*}[t]
\centering
\small
\begin{tabular}{llcrrrrrrrrr}
\toprule
& & & \multicolumn{2}{c}{\textbf{ARC}} & \multicolumn{2}{c}{\textbf{Belebele}} & \multicolumn{2}{c}{\textbf{HellaSwag}} & \multicolumn{2}{c}{\textbf{MMLU}} \\
\cmidrule(lr){4-5} \cmidrule(lr){6-7} \cmidrule(lr){8-9} \cmidrule(lr){10-11}
\textbf{Language} & \textbf{Model} & \textbf{\#Exp} & \textbf{EN} & \textbf{Target} & \textbf{EN} & \textbf{Target} & \textbf{EN} & \textbf{Target} & \textbf{EN} & \textbf{Target} \\
\midrule
\multirow{3}{*}{Turkish} & Dense & - & 51.11 & 33.56 & 74.11 & 59.11 & 76.33 & 42.80 & 56.45 & 42.35 \\
& LayerMoE & 84 & 49.91 & 36.55 & 75.33 & 61.78 & 76.28 & 49.60 & 56.05 & 43.51 \\
& NeuronMoE & 50 (-40\%) & 49.83 & 34.50 & 74.44 & 61.33 & 76.22 & 47.83 & 56.21 & 43.37 \\
\midrule
\multirow{3}{*}{Hungarian} & Dense & - & 51.11 & 34.85 & 74.11 & 59.67 & 76.33 & 44.63 & 56.45 & 43.68 \\
& LayerMoE & 84 & 49.91 & 39.73 & 75.33 & 64.22 & 76.28 & 51.66 & 56.05 & 44.86 \\
& NeuronMoE & 47 (-44\%) & 50.43 & 37.67 & 74.78 & 60.11 & 76.47 & 49.90 & 56.10 & 44.30 \\
\bottomrule
\end{tabular}
\caption{Stage 1 results for Turkish and Hungarian extension on Llama-3.2-3B. NeuronMoE achieves substantial parameter efficiency while successfully extending to additional low-resource languages. All values in \%. Results for LayerMoE baseline are from \citet{zhang-etal-2025-less}.}
\label{tab:additional_langs}
\end{table*}

Table~\ref{tab:allocation} compares layer-wise expert allocation between LayerMoE and NeuronMoE.
LayerMoE employs gradual reduction (5→2→3→4), while NeuronMoE shows substantial reduction in middle layers (average 1.1 experts for layers 3-20) with selective concentration in early (layer 0: 6 experts) and late layers (layers 21, 23-27: 2-4 experts).

Performance varies systematically across benchmark types.
ARC Challenge (commonsense reasoning) exhibits the largest gaps (2.0-2.5\% degradation vs. LayerMoE), while language understanding tasks show smaller differences: Belebele (0.5-2.8\%), HellaSwag (1.8-3.4\%), and MMLU (0.1-0.4\%).
This pattern reflects our allocation strategy that minimizes experts in middle layers performing abstract reasoning while concentrating capacity in early/late layers handling language-specific processing.

\section{Analysis}

To understand the mechanisms underlying neuron-guided allocation's efficiency, we analyze language-specific neuron distributions in trained MoE models.
We examine post-training expert specialization patterns, compare them with source language distributions, and validate the empirical basis for our allocation strategy.

\subsection{Language-Specific Neuron Distribution in MoE Experts}

We analyze language-specific neuron development in Greek experts after MoE extension using Average Precision (AP) scores.

\paragraph{Analysis Method:}
Measuring language-specific neurons in sparse MoE architectures requires addressing the challenge that each expert processes only token subsets determined by routing decisions.
Thus we extend the methodology of \citet{kojima-etal-2024-multilingual} by tracking routing information $R_l = \{\text{expert\_ids}, \text{routing\_weights}\}$ at each layer and aggregating token-level activations to sample-level representations per expert:
\begin{equation}
h_e^{(s)} = \frac{1}{|T_e^{(s)}|} \sum_{t \in T_e^{(s)}} a_e^{(t)}
\end{equation}
where $T_e^{(s)}$ denotes the set of tokens from sample $s$ routed to expert $e$, and $a_e^{(t)}$ represents expert $e$'s activation vector for token $t$.

Following the AP algorithm, we measure how strongly each neuron correlates with specific languages.
For neuron $n$ in expert $e$, we collect its activation values $h_{e,n}^{(1)}, h_{e,n}^{(2)}, \ldots, h_{e,n}^{(N)}$ across all $N$ samples and compute:
\begin{equation}
\text{AP}_{exp}(n, e, \ell) = \text{AP}(h_{e,n}^{(1:N)}, y_{\ell}^{(1:N)})
\end{equation}
where $y_{\ell}^{(i)} \in \{0,1\}$ indicates whether sample $i$ belongs to language $\ell$.
High AP scores approaching 1.0 identify neurons that consistently activate strongly for specific languages.
We define high-AP neurons as those with AP $\geq$ 0.9, representing strong language specialization.
These scores quantify language-specific capacity that develops during MoE training, enabling us to validate whether our allocation strategy correctly identifies layers requiring expert capacity.

\paragraph{Results:}
Table~\ref{tab:expert_specialization} quantifies high-AP neuron counts per expert across selected layers.
In layer 0 (6 experts allocated), expert\_3 contains 61 Greek-specific neurons (0.31\% of layer neurons), expert\_1 contains 27 neurons (0.14\%), while remaining experts show negligible specialization.
In layer 27 (4 experts allocated), expert\_2 exhibits concentrated specialization with 405 Greek-specific neurons (2.08\%), whereas expert\_0 and expert\_1 contain zero Greek-specific neurons.
Middle layers (3-20) with single-expert allocation exhibit minimal language-specific neuron development, with most layers containing fewer than 10 specialized neurons total.

\begin{figure}[t]
\centering
\includegraphics[width=0.48\textwidth]{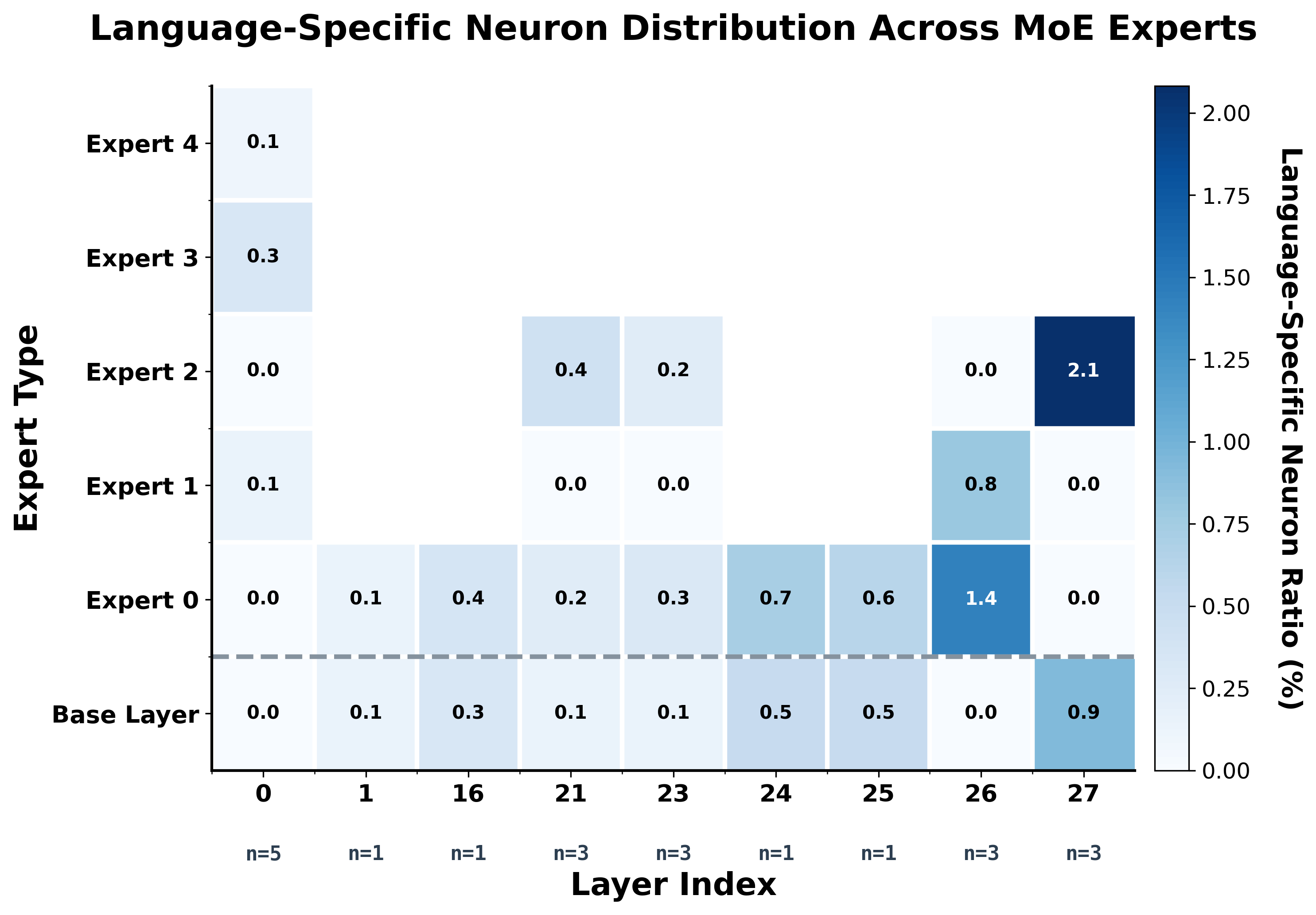}
\caption{Heatmap of high-AP neuron ratios for Greek across all experts. Color intensity indicates Greek specialization strength (red: high, blue: low/none). Multi-expert layers (0, 21, 23-27) show distinct specialization patterns across experts, while single-expert middle layers remain largely unspecialized, validating our neuron-guided allocation strategy.}
\label{fig:expert_heatmap}
\end{figure}

Figure~\ref{fig:expert_heatmap} visualizes this heterogeneous distribution: high-AP neuron ratios concentrate in early layers (0-2) and late layers (21, 23-27), while middle layers show minimal language-specific processing across all experts.

\begin{table}[t]
\centering
\small
\begin{tabular}{lrr}
\toprule
\textbf{Layer-Expert} & \textbf{High-AP} & \textbf{Ratio (\%)} \\
\midrule
L0 base\_layer & 0 & 0.00 \\
L0 expert\_0 & 0 & 0.00 \\
L0 expert\_1 & 27 & 0.14 \\
L0 expert\_2 & 0 & 0.00 \\
L0 expert\_3 & 61 & 0.31 \\
L0 expert\_4 & 19 & 0.10 \\
\midrule
L26 base\_layer & 0 & 0.00 \\
L26 expert\_0 & 278 & 1.43 \\
L26 expert\_1 & 154 & 0.79 \\
L26 expert\_2 & 0 & 0.00 \\
\midrule
L27 base\_layer & 179 & 0.92 \\
L27 expert\_0 & 0 & 0.00 \\
L27 expert\_1 & 0 & 0.00 \\
L27 expert\_2 & 405 & 2.08 \\
\bottomrule
\end{tabular}
\caption{High-AP neuron counts for Greek across experts in selected layers (total neurons per layer/expert: 19,456). Multi-expert layers show distinct specialization patterns, validating allocation decisions.}
\label{tab:expert_specialization}
\end{table}

\subsection{Implications for Allocation Generalization}

The observed convergence provides an explanation for why neuron-guided allocation generalizes across languages and architectures.
Our allocation strategy derives from analyzing source language (English) neuron distributions to determine per-layer expert counts.
Post-training analysis reveals the mechanism behind this convergence: Greek-specific neurons in middle layers remain minimal (Figure~\ref{fig:expert_heatmap} shows base layer and experts contain 0.0-0.4\% ratios for layers 16-23), while late layers develop concentrated Greek specialization (Expert\_0 has 1.4\% in layer 26, Expert\_2 reaches 2.1\% in layer 27).

This pattern demonstrates that convergence occurs not through middle layers adapting to match English distribution, but through target languages developing their own late-layer concentration that complements the existing early-layer focus.
Greek maintains its language-specific characteristics in middle layers while adding substantial capacity in late layers (Table~\ref{tab:lang_comparison}), resulting in a distribution pattern similar to English despite different absolute values and layer-specific peaks.

This mechanism indicates our method exploits the principles: transformer models organize language-specific processing at layer boundaries (early for input encoding, late for output generation) while middle layers handle language-agnostic abstract reasoning.
The strategy concentrates experts where linguistic specialization empirically manifests, explaining the 40\% parameter reduction while maintaining performance across typologically diverse languages.

\section{Discussion}

Our 40-50\% parameter reduction stems from identifying that middle layers require minimal language-specific capacity, enabling single-expert allocation where LayerMoE allocates 2-3 experts.
The performance tradeoffs are task-dependent: ARC Challenge (commonsense reasoning) shows 2.0-2.5\% degradation, while Belebele and MMLU (language understanding) show smaller gaps (0.5-2.8\% and 0.1-0.4\% respectively), and HellaSwag (reading comprehension) shows intermediate degradation (1.8-3.4\%).
This pattern suggests middle layer reduction disproportionately affects commonsense reasoning over language understanding tasks.
The convergence of specialization patterns across typologically diverse languages provides evidence for universal functional organization, implying practitioners can leverage allocation strategies from high-resource languages when extending to new languages.

\section{Conclusion}

We propose neuron-guided MoE allocation for efficient multilingual LLM extension.
Unlike layer-wise similarity-based approaches, we directly measure language-specific neurons across all transformer components to guide expert allocation based on empirical specialization patterns.
Experiments on Llama-3.2-3B (Greek, Turkish, Hungarian) and Qwen-1.5-1.8B achieve 40-50\% parameter reduction with comparable performance.
The key finding is that early and late layers concentrate language-specific neurons requiring multiple experts, while middle layers show minimal specialization enabling single-expert allocation.
Target languages independently develop similar patterns despite typological differences, indicating potential universal architectural principles that explain generalization across languages and architectures.

\section*{Limitations}
Our experiments cover three typologically diverse languages from Indo-European, Turkic, and Uralic families.
While our current study covers typologically diverse languages, extending validation to broader language families such as Indo-Aryan (e.g., Bengali, Nepali), Sino-Tibetan, and Niger-Congo would be a valuable direction for future research.
Regarding the hyperparameters $E_{min}$ and $E_{max}$ used in expert allocation scaling, our ablation study (NeuronMoE-EN) suggests that performance is relatively robust to allocation variations; however, further sensitivity analysis could provide additional insights into optimal hyperparameter selection.
Neuron analysis requires one-time preprocessing to measure language-specific activations, though this cost amortizes across subsequent training runs for the same architecture and language pair.
Finally, since our evaluation is limited to multiple-choice QA tasks, generalization to other tasks is not guaranteed.

\section*{Acknowledgements}
We thank the three anonymous reviewers for their
helpful comments and feedback. 
We also thank Taisei Yamamoto, Xiaotian Wang and Daiki Matsuoka for their insightful discussions.
This work was
supported by JST CREST Grant Number JPMJCR2565, Japan and JST BOOST Program Grant Number JPMJBY24H5, Japan.
\bibliography{cited}

\appendix

\section{Cross-Lingual Neuron Distribution}
\label{sec:neuron_distribution}

This section provides the complete cross-lingual neuron diversity measurements (EN+EL) that inform NeuronMoE's allocation strategy (Table~\ref{tab:neuron_diversity}).
\begin{table}[ht]
\centering
\small
\begin{tabular}{lc}
\toprule
\textbf{Layers} & \textbf{Unique Neurons (EN+EL)} \\
\midrule
0-2 & 342, 79, 11 \\
3-10 & 18, 15, 11, 8, 7, 8, 8, 7 \\
11-18 & 6, 9, 14, 9, 18, 81, 20, 19 \\
19-27 & 28, 38, 232, 51, 208, 114, 111, 238, 223 \\
\midrule
\textbf{Total} & \textbf{1,933} \\
\bottomrule
\end{tabular}
\caption{Cross-lingual neuron diversity across layers (Llama-3.2-3B, Greek extension).}
\label{tab:neuron_diversity}
\end{table}

\section{Cross-Lingual Neuron Distribution Comparison}
\label{sec:lang_comparison}

This section presents the complete layer-by-layer comparison of language-specific neuron counts across English (source) and three target languages (Greek, Turkish, Hungarian) representing three distinct language families (Table~\ref{tab:lang_comparison}).
Despite typological differences, all languages exhibit the same pattern: minimal specialization in middle layers with concentration at layer boundaries, supporting the universal architectural principles discussed in Section 6.

\begin{table}[t]
\centering
\scriptsize
\begin{tabular}{lrrrr|lrrrr}
\toprule
\textbf{L} & \textbf{EN} & \textbf{EL} & \textbf{TR} & \textbf{HU} & \textbf{L} & \textbf{EN} & \textbf{EL} & \textbf{TR} & \textbf{HU} \\
\midrule
0 & 322 & 21 & 8 & 8 & 14 & 7 & 2 & 1 & 1 \\
1 & 73 & 6 & 5 & 3 & 15 & 10 & 8 & 4 & 6 \\
2 & 8 & 3 & 3 & 1 & 16 & 11 & 70 & 10 & 8 \\
3 & 11 & 7 & 17 & 6 & 17 & 4 & 16 & 15 & 8 \\
4 & 13 & 2 & 0 & 1 & 18 & 10 & 9 & 9 & 15 \\
5 & 9 & 2 & 2 & 2 & 19 & 9 & 19 & 22 & 23 \\
6 & 7 & 1 & 0 & 1 & 20 & 16 & 22 & 27 & 18 \\
7 & 6 & 1 & 1 & 0 & 21 & 17 & 215 & 68 & 92 \\
8 & 5 & 3 & 1 & 0 & 22 & 24 & 27 & 47 & 40 \\
9 & 7 & 1 & 1 & 0 & 23 & 70 & 139 & 137 & 117 \\
10 & 7 & 0 & 1 & 1 & 24 & 63 & 51 & 173 & 190 \\
11 & 5 & 1 & 1 & 1 & 25 & 60 & 52 & 185 & 303 \\
12 & 8 & 1 & 0 & 0 & 26 & 75 & 163 & 180 & 124 \\
13 & 9 & 5 & 2 & 0 & 27 & 108 & 118 & 70 & 24 \\
\bottomrule
\end{tabular}
\caption{Complete layer-wise comparison of language-specific neuron counts across English (base Llama-3.2-3B), Greek, Turkish, and Hungarian}
\label{tab:lang_comparison}
\end{table}

\section{Expert Allocation Patterns for Additional Languages}
\label{sec:additionallanguage}

This section provides the complete expert allocation patterns for Turkish and Hungarian extensions (Table~\ref{tab:allocation_additional}), complementing the Greek allocation shown in Table~\ref{tab:allocation}.
These allocations demonstrate that the neuron-guided strategy generalizes across typologically diverse languages from different families.

Table~\ref{tab:allocation_additional} shows the layer-wise expert allocation patterns for Turkish and Hungarian, following the same neuron-guided allocation strategy applied to Greek.
Both languages exhibit similar heterogeneous patterns with concentration in early and late layers, validating the universality of our approach across diverse language families.

The allocation patterns reveal consistent principles across languages: early layers (0-2) receive 5-6 experts for language-specific input processing, middle layers (3-20) predominantly use single experts for language-agnostic reasoning, and late layers (19-27) show varying concentration (2-6 experts) based on language-specific generation requirements.
Despite minor differences in total expert count (47-50, all three languages achieve substantial parameter reduction compared to LayerMoE's relatively uniform 84 experts while maintaining strong performance.

\section{Training Configuration Details}
\label{sec:training}

All MoE extension experiments used identical hyperparameters across languages (Greek, Turkish, Hungarian) and architectures (Llama-3.2-3B, Qwen-1.5-1.8B) to ensure fair comparison.

\paragraph{Training Data:}
Stage 1 uses 2B tokens per target language from CulturaX \citep{nguyen-etal-2024-culturax}.
Stage 2 router training samples 50K examples each from SlimPajama \citep{cerebras2023slimpajama} for English, SkyPile-150B \citep{skypile150b} for Chinese, and CulturaX for Spanish.

\paragraph{Baselines:}
Dense (no MoE), LayerMoE-S1/LayerMoE (84 experts for Llama, 72 for Qwen), and NeuronMoE variants (49 experts for Llama, 36 for Qwen).
NeuronMoE-EN (37 experts) allocates based solely on English neuron distributions.

\paragraph{Computational Cost:}
Neuron analysis requires approximately six GPU minutes per language on a single NVIDIA GH200.
Training uses DeepSpeed ZeRO Stage 2 on NVIDIA GH200 96GB GPUs, requiring approximately 120 GPU hours total.

Complete hyperparameters are shown in Table~\ref{tab:training_config}.

\begin{table}[h]
\centering
\small
\begin{tabular}{ll}
\toprule
\textbf{Parameter} & \textbf{Value} \\
\midrule
\multicolumn{2}{l}{\textit{Optimization:}} \\
Optimizer & AdamW \\
Learning rate & 1e-4 \\
Adam $\beta_1$, $\beta_2$ & 0.9, 0.999 \\
Adam $\epsilon$ & 1e-8 \\
Weight decay & 0.0 \\
LR scheduler & Cosine \\
Warmup steps & 0 \\
\midrule
\multicolumn{2}{l}{\textit{Training:}} \\
Batch size per device & 32 \\
Gradient accumulation steps & 1 \\
Effective batch size & 32 \\
Total training steps & 15,000 \\
Max gradient norm & 1.0 \\
Precision & BF16 \\
Seed & 42 \\
\midrule
\multicolumn{2}{l}{\textit{MoE-specific:}} \\
Top-K routing & 2 \\
Auxiliary loss coefficient & 0.01 \\
Min experts per layer ($E_{\min}$) & 1 \\
Max experts per layer ($E_{\max}$) & 6 \\
\bottomrule
\end{tabular}
\caption{Complete training hyperparameters for all experiments. All models trained with identical settings for fair comparison.}
\label{tab:training_config}
\end{table}

Table~\ref{tab:expert_allocation} shows the detailed per-layer expert allocation for both LayerMoE and NeuronMoE across Llama-3.2-3B and Qwen-1.5-1.8B architectures.

\begin{table}[t]
\centering
\scriptsize
\begin{tabular}{lccc}
\toprule
\textbf{Layers} & \textbf{Greek} & \textbf{Turkish} & \textbf{Hungarian} \\
\midrule
0-2 (Early) & 6,2,1 & 6,2,1 & 5,2,1 \\
3-10 (Early-Mid) & 1$\times$8 & 1$\times$8 & 1$\times$8 \\
11-18 (Mid-Late) & 1$\times$7,2 & 1$\times$8 & 1$\times$8 \\
19-27 (Late) & \begin{tabular}{@{}c@{}}1,1,4,1,4,\\2,2,4,4\end{tabular} & \begin{tabular}{@{}c@{}}1,1,1,2,2,\\4,4,4,4,3\end{tabular} & \begin{tabular}{@{}c@{}}1,1,1,2,1,\\3,4,6,3,2\end{tabular} \\
\midrule
\textbf{Total} & \textbf{49} & \textbf{50} & \textbf{47} \\
\bottomrule
\end{tabular}
\caption{Layer-wise expert allocation for Turkish and Hungarian on Llama-3.2-3B (28 layers). Both languages show similar allocation patterns to Greek, with concentration in input-proximal and output-proximal layers.}
\label{tab:allocation_additional}
\end{table}

\begin{table}[t]
\centering
\small
\begin{tabular}{lp{5.0cm}}
\toprule
\textbf{Method} & \textbf{Expert Allocation (Layer 0-23)} \\
\midrule
LayerMoE & [5, 4, 3, 5, 4, 3, 2, 3, 3, 2, 2, 2, \newline \phantom{[}2, 2, 2, 2, 2, 3, 3, 4, 4, 4, 3, 3] \\
NeuronMoE & [4, 2, 2, 2, 1, 1, 1, 1, 1, 1, 1, 1, \newline \phantom{[}1, 1, 1, 1, 1, 1, 1, 1, 1, 1, 2, 6] \\
\bottomrule
\end{tabular}
\caption{Per-layer expert allocation for LayerMoE and NeuronMoE on Qwen-1.5-1.8B (24 layers). LayerMoE uses attention-layer similarity; NeuronMoE uses cross-lingual neuron diversity analysis.}
\label{tab:expert_allocation}
\end{table}

NeuronMoE's allocation pattern reflects the empirical finding that language-specific neurons concentrate in early and late layers, while middle layers remain largely language-agnostic.
This neuron-guided strategy achieves 41.3\% (Llama) and 62.5\% (Qwen) parameter reduction compared to LayerMoE's attention-based allocation.

\section{Detailed Expert Specialization Statistics}

Table~\ref{tab:expert_stats_full} presents the complete statistics of high-AP neuron counts for Greek across all experts in layers that received multiple experts under our neuron-guided allocation strategy. These measurements were taken after Stage 1 training.
\begin{table}[t]
\centering
\scriptsize
\begin{tabular}{rrrrr}
\toprule
\textbf{L} & \textbf{Expert} & \textbf{Total} & \textbf{High-AP} & \textbf{Ratio (\%)} \\
\midrule
0 & base & 19,456 & 0 & 0.00 \\
0 & exp\_0 & 19,456 & 0 & 0.00 \\
0 & exp\_1 & 19,456 & 27 & 0.14 \\
0 & exp\_2 & 19,456 & 0 & 0.00 \\
0 & exp\_3 & 19,456 & 61 & 0.31 \\
0 & exp\_4 & 19,456 & 19 & 0.10 \\
\midrule
1 & base & 19,456 & 27 & 0.14 \\
1 & exp\_0 & 19,456 & 27 & 0.14 \\
\midrule
16 & base & 19,456 & 64 & 0.33 \\
16 & exp\_0 & 19,456 & 71 & 0.36 \\
\midrule
21 & base & 19,456 & 26 & 0.13 \\
21 & exp\_0 & 19,456 & 46 & 0.24 \\
21 & exp\_1 & 19,456 & 0 & 0.00 \\
21 & exp\_2 & 19,456 & 82 & 0.42 \\
\midrule
23 & base & 19,456 & 26 & 0.13 \\
23 & exp\_0 & 19,456 & 57 & 0.29 \\
23 & exp\_1 & 19,456 & 0 & 0.00 \\
23 & exp\_2 & 19,456 & 46 & 0.24 \\
\midrule
24 & base & 19,456 & 100 & 0.51 \\
24 & exp\_0 & 19,456 & 139 & 0.71 \\
\midrule
25 & base & 19,456 & 98 & 0.50 \\
25 & exp\_0 & 19,456 & 118 & 0.61 \\
\midrule
26 & base & 19,456 & 0 & 0.00 \\
26 & exp\_0 & 19,456 & 278 & 1.43 \\
26 & exp\_1 & 19,456 & 154 & 0.79 \\
26 & exp\_2 & 19,456 & 0 & 0.00 \\
\midrule
27 & base & 19,456 & 179 & 0.92 \\
27 & exp\_0 & 19,456 & 0 & 0.00 \\
27 & exp\_1 & 19,456 & 0 & 0.00 \\
27 & exp\_2 & 19,456 & 405 & 2.08 \\
\bottomrule
\end{tabular}
\caption{Complete expert specialization statistics for Greek. High-AP neurons indicate strong Greek specialization developed during Stage 1 training. Total neurons per layer/expert: 19,456.}
\label{tab:expert_stats_full}
\end{table}

\section{Detailed Results for Additional Languages}
\label{sec:additional_results}

We also provide preliminary Stage 1 (Expert Initialization) results for Bengali (Indo-Aryan) and Nepali (Indo-Aryan) in Table~\ref{tab:stage1_additional}. These results further demonstrate the extensibility of our method to languages beyond the primary experimental set.

\begin{table*}[t]
\centering
\small
\begin{tabular}{llrrrr}
\toprule
\textbf{Model} & \textbf{Language} & \textbf{ARC} & \textbf{Belebele} & \textbf{HellaSwag} & \textbf{MMLU} \\
\midrule
Dense & English & 51.11 & 74.11 & 76.33 & 56.45 \\
& Nepali & 24.12 & 42.44 & 32.54 & 33.77 \\
& Bengali & 26.26 & 44.78 & 33.26 & 33.48 \\
\midrule
NeuronMoE-S1 (NE) & English & 47.44 & 71.44 & 75.29 & 54.71 \\
& Nepali & 24.89 & 44.67 & 33.07 & 34.63 \\
\midrule
NeuronMoE-S1 (BN) & English & 48.72 & 75.33 & 76.01 & 55.28 \\
& Bengali & 26.35 & 46.89 & 33.57 & 33.47 \\
\bottomrule
\end{tabular}
\caption{Stage 1 results for Bengali and Nepali extensions on Llama-3.2-3B.}
\label{tab:stage1_additional}
\end{table*}
\end{document}